# Evidence of coevolution in multi-objective evolutionary algorithms

James M. Whitacre

School of Information Technology and Electrical Engineering; University of New South Wales at the Australian Defence Force Academy, Canberra, Australia

*Abstract -* This paper demonstrates that simple yet important characteristics of coevolution can occur in evolutionary algorithms when only a few conditions are met. We find that interaction-based fitness measurements such as fitness (linear) ranking allow for a form of coevolutionary dynamics that is observed when 1) changes are made in what solutions are able to interact during the ranking process and 2) evolution takes place in a multi-objective environment. This research contributes to the study of simulated evolution in a at least two ways. First, it establishes a broader relationship between coevolution and multi-objective optimization than has been previously considered in the literature. Second, it demonstrates that the preconditions for coevolutionary behavior are weaker than previously thought. In particular, our model indicates that direct cooperation or competition between species is not required for coevolution to take place. Moreover, our experiments provide evidence that environmental perturbations can drive coevolutionary processes; a conclusion that mirrors arguments put forth in dual phase evolution theory. In the discussion, we briefly consider how our results may shed light onto this and other recent theories of evolution.

## 1 Introduction

In Evolutionary Algorithms (EAs), individuals (i.e. candidate solutions to the problem) typically interact in two ways: through selective and recombinative operators. Both forms of interaction impact algorithm behavior by controlling how the population moves through solution space. Recombinative operators (or search operators in general) influence how the population can sample new positions in solution space while selective operators control how, and under what conditions, solutions are added and removed from the population.

Although the two operators are quite distinct, the execution of either operator requires some interaction between individuals in order to determine which new solutions to sample and which to keep. This fact can sometimes be obscured when implementing globally defined selection schemes or when using distribution based search techniques where these interactions are not explicitly defined. Such interactions are more clearly observed however in EAs using tournament selection, EAs with more traditional crossover operators, and in most spatially distributed EA designs such as the cellular GA. In these latter cases, there are clearly pair-wise or small group interactions which result in population-wide search behaviors.

## 1.1 Coevolution

Under some conditions, fitness evaluations involving interactions between individuals can result in a contextual or subjective definition of fitness. By *contextual fitness*, we mean that fitness can sometimes be sensitive to who the individual interacts with during fitness evaluation. Simulation studies have suggested that contextual fitness may influence natural phenomena such as speciation [1], population diversity [2], and cooperative behavior [3].

Contextual fitness is also a key defining feature of coevolution where fitness is determined in part based on interactions between individuals. Coevolution has been studied in a number of computer models in the domain of artificial life (e.g. NKC models [4] [5] and tangled nature models [6]), evolutionary game theory [7] [8], and in optimization (e.g. cooperative [9] and competitive coevolutionary algorithms [10]). With coevolution, the evaluation of fitness can sometimes be intransitive, meaning that changes to the interaction topology (i.e. who interacts with who) can alter the ordering of fitness rankings (e.g. likelihood of survival) of individuals.

**Coevolving Fitness Landscapes:** Descriptions of coevolution often focus entirely on contextual fitness. A similar but broader view describes coevolution based on the contextual dependence of fitness landscapes between interacting species (e.g. see [4]). This latter perspective requires one to look beyond current fitness evaluations and consider the contextual dependence of future adaptive options for each species. In particular, the mutations (to a species) that are deemed adaptive will depend on the context in which mutations are taking place. This paper studies coevolution by evaluating the codependence of fitness landscapes. However, because this is a direct consequence of contextual fitness, this should not detract from any of the conclusions drawn here.

## 1.2 Multi-Objective Optimization

Similar to coevolution, Multi-Objective Optimization (MOO) involves a class of problems where a solution's fitness is contextually dependent on who the solution interacts with during fitness evaluation. For example, consider a fitness function defined based on linear ranking, e.g. where pair-wise comparisons are used to count the number of solutions in a population that a particular solution dominates. The fitness function is shown in (1) where the population size is $\mu$. Here we use the standard connotation of the term dominance, where one solution $x_i$ dominates another $x_j$ if the solution is better than another in at least one objective and is equal or better in all other objectives [11].

$$F(x_i) = rank(x_i) = \sum_{j \neq i, j \in \mu} dom(x_i, x_j) \qquad (1)$$

Like coevolution, there are situations in MOO where the relative viability (e.g. ranking order) of two solutions can flip depending on the other solutions in the population. This is illustrated in Figure 1 where two solutions, X and Y, are being compared to a third solution Z based on two objectives, $Obj_1$ and $Obj_2$. In Figure 1a, solution Z is only dominated by Y, while in Figure 1b solution Z is only dominated by X. From this simple illustration, one can see that dominance ranking in MOO can be sensitive to the context (e.g. the definition of Z). From hereafter, we define strong/weak contextual sensitivity of a fitness evaluation based on the occurrence of different contexts that can/cannot change the rankings of individuals.

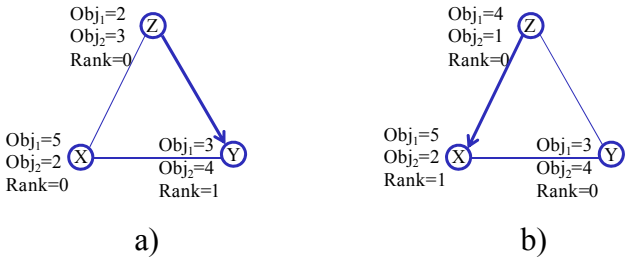

a)          b)

Figure 1: Contextual dependence of dominance ranking. Dominance is indicated by directed connections.

Although Figure 1 illustrates how a strong contextual sensitivity is *possible* in MOO, it does not tell us whether strong sensitivity is commonly observed or whether it is relevant to the dynamics of a multi-objective EA (MOEA) in practice. If the contextual sensitivity is weak in MOO, then changing the context for fitness evaluation will not morph the fitness landscape enough to change the adaptive options available to a solution's offspring. For example, consider the case where a solution stabilizes (converges in objective space). If contextual sensitivity is weak, then changes to solution interactions during fitness evaluation will not be able to change this stability. On the other hand, if contextual sensitivity is strong, changes in context may sometimes destabilize an otherwise stable solution and allow for new adaptations to take place. It is this second case that is frequently observed in biological evolution and that would be interesting to observe in MOO.

Previous studies have also investigated the relationship between multi-objective optimization and coevolution [12] [13] [8]. In [12], they describe how coevolution can be framed as a multi-objective optimization problem where every individual used to evaluate the fitness of another can be thought of as a distinct objective for that individual. They go on to argue that multi-objective optimization techniques and concepts such as Pareto dominance can be directly applied to coevolutionary systems which they later demonstrate in [13] (but also see [14]). Furthermore, concepts such as Pareto dominance have been found to be useful in guaranteeing monotonic improvements in coevolutionary systems [15]. However, unlike these previous studies which have focused on applying concepts from multi-objective optimization to coevolution, this paper investigates whether coevolution is an observable feature in standard MOEAs.

## 1.3 Outline

This study investigates whether population-based search algorithms operating in an MOO environment can display characteristics commonly attributed to coevolution. To test this, it is necessary to design experimental conditions so that the ONLY interaction between individuals in an evolutionary algorithm is through fitness ranking. The next section describes the main features of such an EA which is followed by experimental results in Section 3. Section 4 discusses how our results may be important to two competing theories of evolution: Self-Organized Criticality and Dual Phase Evolution. Conclusions are drawn in Section 5.

## 2 Experimental Setup

### 2.1 Species Model

Several experimental conditions must be observed in order to investigate whether interaction-based fitness measurements such as linear ranking can enable coevolutionary behavior in an EA. The pseudocode for our algorithm highlights these changes and is provided in Figure 3.

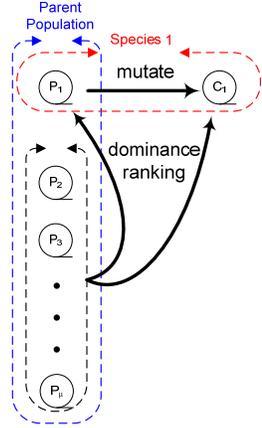

Figure 2: fitness evaluation and species adaptation in ESIM. $C_1$ replaces $P_1$ if its ranking is higher and it is not dominated by $P_1$.

In these experiments, we use an Evolutionary Species Island Model (ESIM) that captures two important features of natural evolution: i) no interbreeding is allowed between species and ii) a constrained set of species participate in defining the fitness of others. In ESIM, each individual represents a distinct species and the adaptation of species occurs through mutation. Also, species only interact through fitness calculations, which involves dominance ranking as defined in (1).

The ESIM does not allow direct competition between species meaning that, although a species can adapt (through mutation), no speciation or extinction events take place. Figure 2 illustrates these aspects of the algorithm. The removal of direct competition between species was viewed as necessary to ensure that the only interaction taking place was through definitions of fitness and not through competitive selection. Also, allowing for competitive selection introduces a branching process into the (species) population dynamics which would make it more difficult to determine the presence of coevolution in our analysis.

**Investigating Coevolution**: Our study of coevolutionary behavior involves changing the context for fitness evaluation (described here) and analyzing whether this can influence the evolution of species (described in Section 3.1). Changes in context occur in the following manner. First, species are grouped into islands where species within the island participate in defining each other's fitness. Migration events between islands are then used as a tool for manually introducing changes to who an individual species interacts with. Migration takes place using a series of pair-wise swaps with the number of swaps controlling the size of the migration event.

Migrations cause changes in which species interact during ranking calculations (defined in (1)). Because each species evolves through a localized sampling of genotype space (and not through recombination or speciation), any changes to the adaptive options of a species after a migration must be a result of morphological changes to the species fitness landscape. Hence, these migration events provide a well-controlled change in context that can be used as a tool for investigating the presence of coevolution.

```
Initialize EA into four equally sized islands
Evaluate each individual (in each island)
Determine Fitness of each individual using (1)
FOR each generation
    FOR each parent in an island
        Generate child C_i by mutating parent P_i
        Evaluate C_i on test problem
        Calculate Fitness F of  P_i ∪ (P- P_i)†
        Calculate Fitness F of  C_i ∪ (P- P_i)†

    FOR each parent in island
        IF  {F(C_i) > F(P_i)} ∧ ¬ {dom(P_i, C_i)}  THEN
C_i replaces P_i

Store phenotypes of new population in archive
    Calculate ΔPh over last τ generations using (2)
Every 300 generations, conduct a migration event
† Ranking takes place between individuals within the same island
```

Figure 3: Pseudocode for ESIM.

**Remaining algorithm details**: The total population size was set to $\mu$=64 with species divided into four equally sized islands (i.e. islands with population size of 16). For each test problem considered, the solution space was defined using real gene encoding of dimensionality $n$. Offspring were copies of a parent but with each gene mutated with a $2/n$ probability and with mutations occurring by locally sampling alleles. Single objective and multi-objective test problems used in these experiments are listed in Table 1. Only two objectives are considered ($m$=2) in MOP experiments. Unless stated otherwise, results are averaged over 30 runs.

Table 1: Single and multiple objective test problems. More information on parameter settings (second and fifth columns) is available in the references provided.

| MOPs | | | SOPs | | |
|---|---|---|---|---|---|
| name | n/m | ref | name | n | ref |
| DTLZ1 | 20/2 | [16] | ECC | n=MN, M=24, N=12 | [17] |
| DTLZ2 | 20/2 | [16] | Griewangk | n=10 | [18] |
| DTLZ3 | 20/2 | [16] | FM | n=6 | [18] |
| DTLZ4 | 20/2 | [16] | Hyper-ellipsoid | n=30 | [19] |
| ZDT3 | 30/2 | [20] | MMDP | n=6k (k=20) | [21] |
| ZDT4 | 10/2 | [20] | MTTP | n=200 | [17] |

# 3 Results

## 3.1 Measuring Fitness Landscape Morphing

To evaluate whether coevolution takes place in the species model, it must be demonstrated that changes in context (for fitness evaluation) can morph the fitness landscape enough to change which mutations are considered positive adaptations. In other words, the order of fitness rankings between the un-mutated and mutated versions of the species (also referred to as the parent and child) must be shown to depend upon the context.

**Measurement Procedure:** Measuring changes to the fitness landscape takes place using the following procedure. During a migration event, a record is kept of each solution's original context (individuals in its island prior to migration) and its new context (individuals in its island after migration). The fitness landscape for a solution is then sampled by generating 100 children using the mutation operator defined previously. Each child is evaluated in the two contexts and we determine (based on the selection procedure in Figure 3) if the child would replace the parent species for each context. We next define the contextual sensitivity as the proportion of children where the change in context results in different replacement outcomes. In other words, the contextual sensitivity for a species is measured as the proportion of reachable points in fitness landscape (i.e. children) that are strongly sensitive to the change in context. The average sensitivity over all species is reported as the sensitivity metric $S$.

A positive value for $S$ indicates that the contextual sensitivity is strong enough to morph the fitness landscape in ways that alter which mutations are positive adaptations. Because such morphing can influence the evolutionary path that a species takes, positive $S$ values provide empirical evidence that coevolution is feasible within the given experimental conditions. Results for the contextual sensitivity metric $S$ are given in Table 2 for both single objective and multiple objective test problems. Here, the contextual sensitivity is found to be strong (positive) only when evolution occurs in a multi-objective environment.

Table 2 Average sensitivity metric S for ESIM on SOP and MOP problems

| SOP | S | MOP | S |
|---|---|---|---|
| ECC | 0.0000 | DTLZ1 | 0.0298 |
| Griewangk | 0.0000 | DTLZ2 | 0.0597 |
| FM | 0.0000 | DTLZ3 | 0.1569 |
| Hyper-ellipsoid | 0.0000 | DTLZ4 | 0.0291 |
| MMDP | 0.0000 | ZDT3 | 0.1018 |
| MTTP | 0.0000 | ZDT4 | 0.0168 |

## 3.2 Observing Coevolutionary Dynamics

Although a positive $S$ indicates that a change in context is strong enough to matter, it does not tell us if the change in context actually influences the *evolutionary path* (sequence of adaptive steps across the fitness landscape) that a species takes. Whether the path that evolution takes is affected in practice depends on the magnitude of $S$ and the attractor characteristics for a particular algorithm searching a particular fitness landscape. Although a rigorous assessment of this behavior is challenging for several reasons, indirect evidence of coevolution can be provided in a relatively straightforward manner.

In preliminary experiments, we found that each species in ESIM tends to converge over time in objective space; a common feature of optimization in a static fitness landscape. Convergence indicates that the accessibility of adaptive steps within the local fitness landscape for each species is decaying over time. If migration events result in brief periods of new adaptive progress, it would indicate that the evolutionary paths of species were in fact being influenced by morphological changes to the fitness landscape. In particular, it would demonstrate that a stable species had become destabilized as a result of changes in interacting species. Here we use this knowledge to provide evidence of coevolutionary dynamics actually occurring in ESIM.

**Measurement Procedure:** The *evolutionary activity* of a species is defined as the movement of a species in objective space over time and is calculated in the following manner. First, the history of phenotypes (objective function values) for each species is recorded in an archive for each generation, as indicated in Figure 3. For a given number of generations $\tau$ over which movement in objective space is being measured, the metric for evolutionary activity $\Delta Ph$ is then defined in (2) as the average change in all $m$ objective functions from all $\mu$ species in the population. For these experiments, $\tau=10$ generations however results were largely insensitive to this parameter.

$$\Delta Ph(t) = \frac{1}{\mu * m} \sum_{i}^{\mu} \sum_{j}^{m} \left| Ph_{i,j,t} - Ph_{i,j,t-\tau} \right| \quad (2)$$

The results in Figure 4 display the time series of evolutionary activity for single objective and multiple objective problems. For visualization purposes, $\Delta Ph$ outputs are rescaled so that each experiment spans over three orders of magnitude and is shifted vertically so that multiple results can be clearly viewed on the same graph.

In all experiments, evolutionary activity decays over time indicating that species are converging towards attractors in their respective fitness landscapes. However during migration events, occurring every 300 generations, we find brief spikes in activity for each of the multi-objective problems. This demonstrates that changes in who a species interacts with in a multi-objective environment can influence the adaptive options taken (i.e. the evolutionary trajectory) by that species. In particular, these interaction changes are able to induce new positive adaptations that would not have taken place otherwise. To assess the significance of this effect, we use a non-parametric statistical test (Mann-Whitney U Test) and calculate the confidence that a distribution of $\Delta Ph$ outputs (at time $t$) has a median value that is greater than the distribution at the time $t-1$. For all of the MOPs (but none of the SOPs) we find that at least one of the migration events has a significant increase in median activity ($p<0.01$). On the other hand, there is never a statistically significant increase in activity at other (non-migration) positions in the time series.

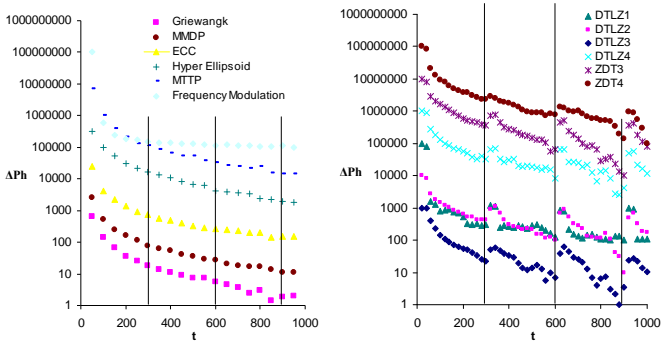

Figure 4: Time series of evolutionary activity for ESIM evolving on a set of single objective (left) and multi-objective (right) problems with large migration events (swaps=100) at every 300 generations.

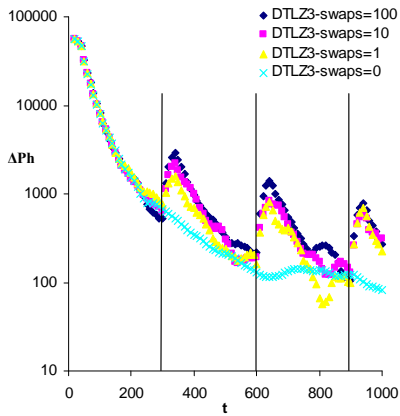

Figure 6 Time series of evolutionary activity for ESIM evolving on DTLZ3 with different migration sizes (swaps=0, 1, 10, 100) which occur at every 300 generations. Results are averaged over 50 runs.

The results in Figure 6 look at the influence of the migration size on evolutionary activity. Results are only shown for DTLZ3 however similar behavior occurs for the other MOPs in

Table 1. This figure confirms that when migration events are prevented between islands in MOO, the species converge without the periodic spikes in evolutionary activity. This figure also demonstrates that the migration size has very little impact and that even the smallest migrations can cause periods of increased evolutionary activity. From this, we speculate that migration events are initiating a cascade of activity such that large portions of the population can be driven to reorganize into new meta-stable states even when the initial size of the migration event is small.

## 4 Discussion

### 4.1 Coevolution in multi-objective optimization

Coevolution describes a phenomena where species influence the fitness landscapes and consequently the evolutionary trajectory of other species. The more general phenomena of coupled dynamics, in which coevolution is a special case, occurs in many contexts and is commonplace in driven systems of interacting components. For instance, an EA population is typically able to converge to a single attractor (or attractor basin) within a single objective fitness landscape as a consequence of selective and recombinative operators. On the other hand, coevolution involves a different form of coupled dynamics where individual species are not driven to the same attractors but instead mutually influence each other's uniquely defined fitness landscapes, which in turn can drive the evolution of individual species in unique directions. Multi-objective problems provide an environment for optimization where individuals are also driven to occupy unique phenotypes (e.g. along the Pareto Front). We suspect that this aspect of multi-objective problems could be directly responsible for the intransitivity of Pareto dominance as well as the coevolutionary behaviors reported here.

**Coevolution through Constraints:** ESIM creates a coevolutionary process that operates by constraining the future adaptive options within a fitness landscape. In particular, the ranking interactions used to define fitness in ESIM act by constraining which movements in objective space are observed as fitness improvements, which movements are neutral and which are negative. This is dissimilar from conventional models of coevolution that are based on explicit acts of cooperation or competition. In future studies, we intend to explore whether "constraints-based" coevolutionary models provide unique opportunities for understanding coevolution.

**Coevolution with single objectives**: Although our results did not find coevolutionary dynamics in SOO, this does not mean that such a phenomenon is not possible. In [2] it is shown how fitness ranking calculations that are restricted to occur through sparsely connected interaction networks can be used to generate a contextual definition of fitness that allows two connected individuals to flip their ranking order when changes take place in the interaction topology. In fact, the study in [2] demonstrates that intransitivity is at least feasible in SOO when using a topologically constrained fitness metric. Future studies will investigate whether the self-organizing topology evolutionary algorithms studied in [2] can also display coevolutionary behavior similar to that demonstrated here in multi-objective environments.

### 4.2 Theories of Evolution

The origin of punctuated dynamics in natural evolution has been hotly debated over the years with much attention given to the theory of Self-Organized Criticality (SOC) [22] [23]. SOC theory attempts to explain how the spatial and temporal patterns in some distributed systems can spontaneously evolve such that changes in the system can reach any size with non-negligible probability. In particular, the theory claims that some coupled dynamical systems are driven or attracted to a critical state where the system displays self-similarity in both space and time as is commonly indicated by power law relations. SOC behavior is different from other critical phenomena (e.g. phase transitions) where an environmental parameter (e.g. temperature) must be tuned in order for the system to reach a critical state.

Power law relations are also displayed in the spatial and temporal properties of natural evolutionary dynamics including extinction sizes, species lifetimes (although this is debated), and taxonomic structure [24] [25], leading some to speculate that natural evolution may be an example of an SOC system where initially small perturbations can grow to large extinction events [26]. Others, notably Kauffman, have argued that critical phenomena could play other roles in evolution [27] [28].

More recently, David Green's Dual Phase Evolution theory (DPE) has argued that the punctuated dynamics in natural evolution are a consequence of external environmental factors (e.g. changes to the physical landscape's connectivity) that can facilitate exploration during the evolution of individual species and can also enable speciation events [29] [30] [31]. In particular, DPE describes two phases for evolution; one involving a highly connected phase where exploitive selective pressure predominates and the overall system is relatively stable, and the other a disconnected phase where variation and exploration predominates. The disconnected phase is initiated by disruptions to the ecosystem but afterwards, the ecosystem reorganizes, often times with greater complexity than it had previously [31].

The experiments in this paper also appear to result in two distinct phases of evolutionary behavior: one where the system converges /stabilizes and another where disruptions (changes in species interaction topology) lead to short bursts of new evolutionary activity (e.g. see Figure 4). Based on this interpretation, the results of our study seem to support the argument that external events can cause meta-stability in evolving populations.

In our experiments, external perturbations come in the form of migration events however the end effect is a change in population connectivity which is arguably similar to changes in a physical landscape as is used in most DPE studies (e.g. see [32]). On the other hand, most supporting evidence for DPE seems to suggest (if not demand) that external perturbations be destructive (e.g. by mass extinctions or redistribution of resources). This is not supported by our results, which instead suggests that changes in connectivity alone might be sufficient in models of coevolution. Such a statement can be clearly and unambiguously made due to our removal of direct competition in the EA. Notice that if we had allowed direct competition to exist in the algorithm, then the observed punctuated dynamics would have been off-handedly attributed to genetic takeover by a newly migrated species. Such genetic takeover would be viewed as a highly disruptive and destructive process that allows for the punctuated dynamics to occur and would be in strong agreement with current explanations of DPE theory.

## 5 Conclusions

In this paper, we have shown that multi-objective evolutionary algorithms can exhibit several features commonly labeled as coevolutionary behavior. We demonstrate that this occurs in a species model when i) individual species participate in defining the fitness of others in a multi-objective environment and ii) when the system is exposed to external perturbations (migrations between islands). We have also explained how these results appears to support a recent theory of evolution known as Dual Phase Evolution Theory, which claims that many features of natural evolution including punctuated dynamics, speciation, and increasing organizational complexity can be partly attributed to events that are initiated by the external environment.